\documentclass[
]{ceurart}

\usepackage{listings}

\begin{document}



\copyrightyear{2025}
\copyrightclause{Copyright for this paper by its authors. Use permitted under Creative Commons License Attribution 4.0 International (CC BY 4.0).}
\conference{ISWC 2025 Companion Volume, November 2--6, 2025, Nara, Japan}

\title{TIB AIssistant: a Platform for AI-Supported Research Across Research Life Cycles}

\author[1]{Allard Oelen}[%
orcid=0000-0001-9924-9153,
email=allard.oelen@tib.eu,
]
\address[1]{TIB – Leibniz Information Centre for Science and Technology, Hannover, Germany}

\author[1,2]{S\"oren Auer}[%
orcid=0000-0002-0698-2864,
email=auer@tib.eu,
]

\address[2]{L3S Research Center, Leibniz University of Hannover, Hannover, Germany}

\begin{abstract}
The rapidly growing popularity of adopting Artificial Intelligence (AI), and specifically Large Language Models (LLMs), is having a widespread impact throughout society, including the academic domain. AI-supported research has the potential to support researchers with tasks across the entire research life cycle. In this work, we demonstrate the TIB AIssistant, an AI-supported research platform providing support throughout the research life cycle. The AIssistant consists of a collection of assistants, each responsible for a specific research task. In addition, tools are provided to give access to external scholarly services. Generated data is stored in the assets and can be exported as an RO-Crate bundle to provide transparency and enhance reproducibility of the research project. We demonstrate the AIssistant's main functionalities by means of a sequential walk-through of assistants, interacting with each other to generate sections for a draft research paper. In the end, with the AIssistant, we lay the foundation for a larger agenda of providing a community-maintained platform for AI-supported research. 
\end{abstract}

\begin{keywords}
  AI Assistant \sep
  AI-Assisted Research \sep
  Scholarly Assistant \sep
  Scholarly AI Platform 
\end{keywords}

\maketitle

\section{Introduction}
\label{sec:introduction}

The recent advancements of Artificial Intelligence (AI), in particular generative AI, such as Large Language Models (LLMs), have a profound impact on our society already~\cite{bommasani2021opportunities}. Among other fields, scholarly research is one of the areas where AI already has, and will likely even more in the future, change how work is conducted~\cite{doi:10.1073/pnas.2314021121}. However, to fully leverage AI for research, the researcher needs to be aware of the many different approaches and needs to have experience with guiding the LLM in such a way that it produces outputs that are useful for their research. Especially due to the growing number of tools, techniques, and approaches, it can be overwhelming and challenging to effectively leverage AI for scholarly research. 

To address these challenges, we present the TIB AIssistant (i.e., AI-assistant). The TIB AIssistant is a domain-agnostic AI-supported research platform that helps researchers, by means of AI, in various steps of the research life cycle. This work demonstrates a concrete implementation of our vision of the TIB AIssistant, which is published as a vision paper~\cite{aissistantvision}. While the vision paper outlines a high-level list of design principles and considerations, this article provides additional value by introducing a concrete implementation of the AIssistant platform, a workflow of different assistants, and a demonstration of the user interactions with the platform. We aim to provide support across the research life cycle by means of different agents (hereafter called \textit{assistants}). This ranges from providing guidance during ideation, helping to share research questions, writing up articles, to publishing a paper. The assistants take in certain inputs and generate specific outputs. If the assistants are used in sequence, generated output is stored and are used as input for the next assistant. However, it is also possible to use the assistants in isolation, i.e., to perform a single task. 
Assistants can use a predefined set of tools in any arbitrary order to accomplish their tasks. Most of these tools are external services, called via REST endpoints, and are called when deemed necessary by the LLM. For example, this makes it possible for assistants to find related work via actual scholarly search platforms, or to find articles based on certain identifiers (e.g., DOIs or ORCIDs). Transparency is a crucial aspect of AI-supported research, meaning that provenance data needs to be published alongside the research output to ensure full openness about the approach. We implement the RO-Crate~\cite{doi:10.3233/DS-210053} approach and use additional ontologies to record how knowledge is created and by whom.

In the end, the AIssistant serves as a library of assistants and tools for academic research, and provides a single platform in which they are integrated. In this work, we demonstrate how the platform operates and show how a researcher can use various assistants, starting from a research idea formulation and fetching related work, to drafting paper sections.

\begin{figure}[t]
    \centering
    \includegraphics[width=0.9\textwidth]{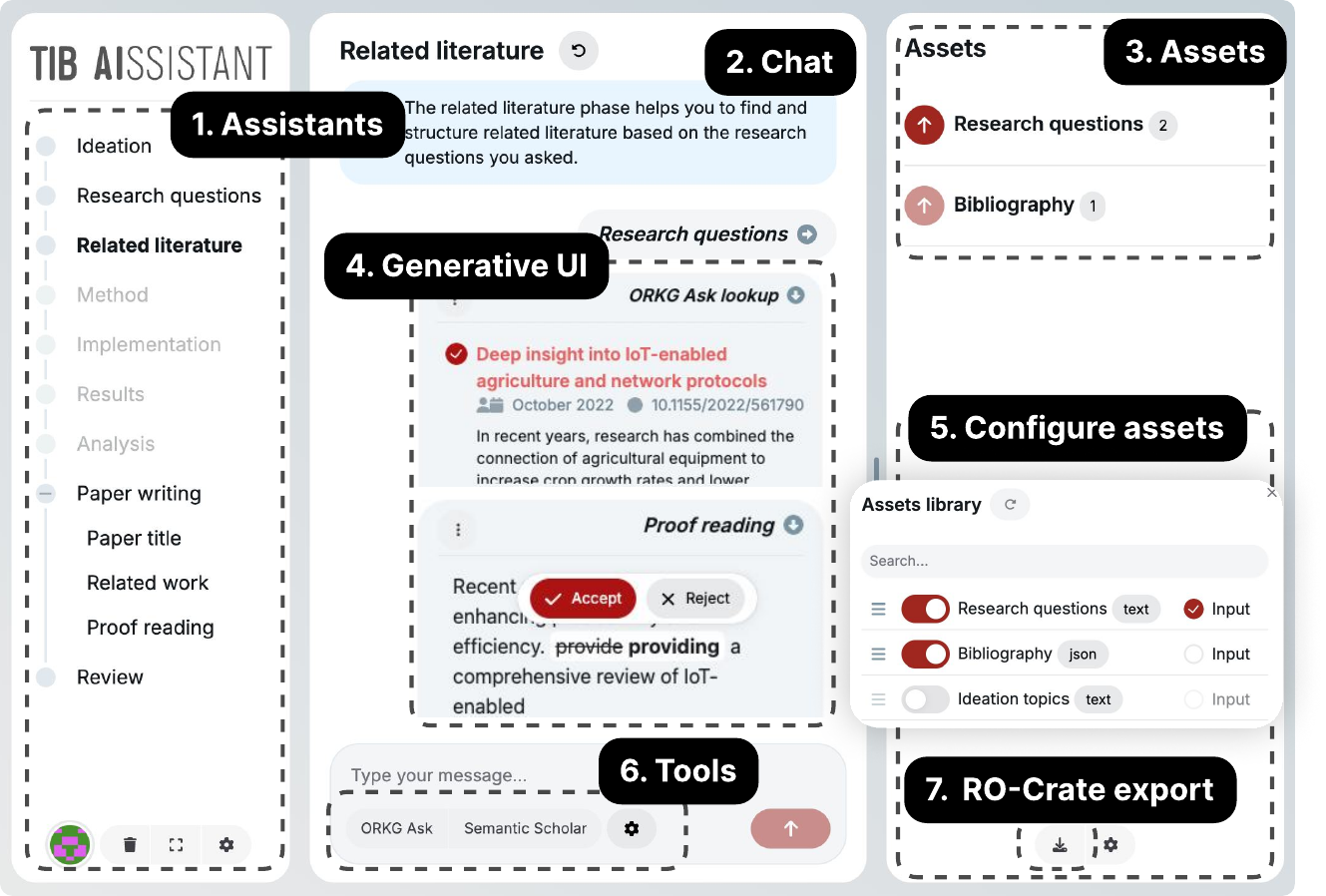}
    \caption{Screenshot of the TIB AIssistant interface. 1. The assistants that can either be used sequentially or individually. 2. Chat window with user and assistant messages, including tool responses. 3. Assets related to the selected assistant. 4. Generative UI selected based on invoked tools (minified example showing two tools: literature search and proofreading). 5. Configure assets; a similar modal is available for assistants and tools. 6. Tools related to the selected assistant. 7. Export assets button to generate a RO-Crate bundle.}
    \label{fig:screenshot}
\end{figure}

\section{Related Work}
\label{sec:related-work}

The industrial revolution and afterward the digital revolution had a substantial impact on all aspects of life~\cite{xu2018fourth}. The rise of AI is considered to be potentially even more impactful than the previous revolutions~\cite{MAKRIDAKIS201746}. Now with the rise of Generative AI, and specifically Large Language Models growing in size, this impact of AI is rapidly accelerated~\cite{tamkin2021understanding}. For example, LLMs are expected to have a significant impact on the labor market~\cite{eloundou2023gpts}. Furthermore, LLMs have an impact on many other fields, including medicine~\cite{clusmann2023future}, finance~\cite{10.1145/3604237.3626869}, and law~\cite{chen2024survey}. LLMs have a significant impact on different aspects of scholarly research as well, for example, for generating research ideas~\cite{si2024can}, literature reviews~\cite{bolanos2024artificial}, and paper writing~\cite{liang2024mapping}. To effectively use LLMs for such purposes, prompt engineering is of crucial importance~\cite{white2023prompt}. The ability to write effective prompts becomes an important skill for professionals seeking to leverage AI for their work~\cite{mesko2023prompt}. To this end, with the TIB AIssistant, we provide a library of prompts (i.e., the assistants) that have been manually curated and can be integrated directly inside research workflows. 

There are many individual approaches to leveraging LLMs for research. Some of these approaches aim to create a fully automated research life cycle using AI. This includes the work of Sakana AI by means of the AI Scientist, which fully automates the research life cycle by iteratively executing life cycle phases~\cite{lu2024ai}. There are several shortcomings of their approach, for example, the limited ability to find related literature, and issues during code generation leading to failing experiments~\cite{beel2025evaluating}. While fully automated approaches are an interesting research avenue and are surely going to be improved in the near future, we take a different approach with the AIssistant. We believe that researchers must be at the center of doing research, and therefore, AI should be an assistive technology, leaving the human researcher in control of the process. 

\section{Approach}
\label{sec:approach}

The AIssistant's approach consists of three main components: assistants, tools, and assets. We will discuss those concepts on a high level to lay the foundation for the platform.

\textbf{Assistants} are agents that are designed to perform a single task. For example, formulating research questions or coming up with a paper title. Assistants consist of a name, system prompt, LLM model, what assets they consume and generate, and the tools that can be called. The LLM model selection determines which model to use to accomplish the task. Simpler tasks can be performed by smaller models, reducing the required resources and waiting time. 

\textbf{Tools} are generally external web services that are called via REST endpoints. The tools have a name, a textual description, an input schema, and an execution function. The description and input schema are provided to the LLM to provide instructions on when the tool should be called and how it should be called. This is handled via the \textit{Tool Calling} mechanism~\cite{schick2023toolformer}. Each assistant specifies the tools that are callable, which limits the space of possible tools and thus improves the tool selection accuracy. The output of tools is either sent to the chat as text or handled by displaying specialized UI components. This follows the paradigm of \textit{Generative UI}, where UI components are shown on demand, depending on the task at hand. In~\autoref{sec:implementation-and-demonstration}, we demonstrate several implemented generative UI components.

\textbf{Assets} are the data store that assistants can access. This differs from the conversation's history (i.e., context window) as it is structured and only contains data that the user selects. Assets have a name and a data type (e.g., text, object, JSON, etc.). Assistants can have required input assets, which either consist of previously generated assets or are manually provided by the user. Reproducibility is one of the cornerstones of science. Therefore, being transparent about how AI was used to support the researcher's work is a crucial aspect of our approach. The assets are central in our approach, as they serve as output of the generated knowledge. We provide an \textit{Export assets} functionality, which compresses assets into an archive file, containing individual files for each asset, and a metadata file which provides provenance data for the files. We use the RO-Crate~\cite{doi:10.3233/DS-210053} for the metadata, which uses JSON-LD and Schema.org annotations to package research artifacts. During the export process, the user can provide additional provenance data (i.e., their name and the license of the data). In addition, we use SPAR ontologies, such as DOCO and DEO~\cite{10.1007/978-3-030-00668-6_8}, to provide more fine-grained semantic annotations for exported data where appropriate. We envision that the generated RO-Crate bundles will be published alongside the research articles to provide transparency of the process. The semantic representations of the generated knowledge provide a more machine-actionable way to access and analyze the knowledge. 

\section{Implementation and Demonstration}
\label{sec:implementation-and-demonstration}
A screenshot of the prototype user interface is displayed in~\autoref{fig:screenshot}. The previously discussed key components are depicted in this screenshot. The three main components, Assistants, Assets, and Tools, are configurable by users. By default, the selected assistant decides what assets and tools are activated, but users can modify this selection to fit their use cases. We will demonstrate here via a system walk-through how a research life cycle using the AIssistant can look like. Afterward, we discuss the technical details of our prototype. 

\subsection{System Walk-Through}

In our walk-through, we adopt the stance of a researcher who is interested in intertwining Semantic Web and AI research. We will follow the life cycle as depicted in the sidebar of~\autoref{fig:screenshot}. For demonstration purposes, we only show domain-agnostic assistants, domain-specific ones are disabled, as can be seen in the screenshot. A list of assistants is provided in \autoref{tab:walkthrough-table}. A demonstration video is available online.\footnote{\url{https://doi.org/10.5446/71179}}

Firstly, the researcher wants to generate ideas related to their topic of interest in the \textit{Ideation} assistant. This assistant asks the user to provide content they are interested in. In our example, the researcher provides a DOI from related papers and an ORCID, which will respectively fetch related content from Crossref and list articles from ORCID. After refining the ideas via the chat, the researcher adds the final topics to the ideation asset. Next, the researcher goes to the \textit{Research questions} assistant and inputs the previously generated ideation topics from the assets. After iterative refinement in the chat component, the final research questions are added to the assets. 

As a next step, the researcher aims to find related work via the \textit{Related literature} assistant. There, the research questions are provided to the chat and are used to find related work via ORKG Ask~\cite{oelen2024orkg}. The user manually reviews the suggested literature and adds relevant articles to the bibliography. This is done via the generative UI approach, where a dedicated literature search component is displayed in the chat window. This component is displayed only for tools that support literature search. It lists articles, allows users to select articles, and provides pagination to explore the listed items further. 

Afterward, the paper writing assistants are used: \textit{Paper title}, \textit{Related work}, and \textit{Proofread}. The first two assistants use the ideation topics and research questions. The related work assistant also uses the bibliography to connect related work to their citations. The final text and bibliography can be exported to LaTeX and imported to an Overleaf project as a draft for writing the actual manuscript. The proofread assistant uses another type of generative UI component, providing a dedicated interface where users can accept and reject proposed changes via a track-changes-like interface. As a final step of the walk-through, the authored text is reviewed via the \textit{Review} assistant.

\begin{table}[]
\centering
\caption{Implemented domain-agnostic assistants following a sequential pipeline. For each assistant, the input and output assets and tools are listed. Certain assistants use input assets from previously generated output assets.}
\label{tab:walkthrough-table}
\resizebox{1\textwidth}{!}{%
\begin{tabular}{@{}p{2.3cm}p{4cm}p{2.7cm}p{3cm}p{3cm}p{2.5cm}@{}}
\toprule
\textbf{Name} & \textbf{Description} & \textbf{Input assets} & \textbf{Output assets} & \textbf{Tools} & \textbf{Generative UI} \\ \midrule
 Ideation & Come up with research ideas based on user-provided input & - & Ideation topics & Crossref, ORCID, PDF URL, Unpaywall & - \\
 Research  questions & Define research questions based on research topics & Ideation topics & Research questions & - &  - \\
 Related literature & Find related literature based on research questions & Research questions & Bibliography & ORKG Ask, Semantic Scholar & Literature search component\\
Paper writing: title & Write a suitable paper title based on provided content & Ideation topics, research questions & Paper: title & - & - \\
Paper writing:  related work & Write a related work section using a bibliography & Research questions, bibliography & Paper: related work & - & - \\
 Paper writing:  proofreading & Proofread the text and propose changes & Paper & Paper &  -  & Track changes component \\
 Review & Review the paper content similarly as a peer reviewer & Paper & Paper & - & - \\
\bottomrule
\end{tabular}%
}
\end{table}

\subsection{Technologies}

The AIssistant is implemented in Typescript and React using the Next.js framework and is published as open source software under a permissive MIT license.\footnote{\url{https://gitlab.com/TIBHannover/orkg/tib-aissistant/web-app}} We use Next.js both as frontend and backend, with the backend primarily serving as a wrapper for making calls to the LLMs. We use the OpenAI API, and specifically the GPT‑4o mini model, which provides a good balance between costs and performance. The Vercel AI SDK ensures that model providers and individual models can be changed when necessary. Local browser storage via IndexedDB is used to store assets and chat history. For authentication, we use ORKG accounts with enable single sign-on. The service can only be used when the user is signed in to ensure the number of daily used tokens can be effectively limited as a means of cost management. In the future, we will implement a Bring Your Own Key (BYOK) approach, allowing users to utilize their own API keys to access the platform without imposed token limits. To encourage community development of both the platform and content, we have published development documentation online.\footnote{\url{https://tibhannover.gitlab.io/orkg/tib-aissistant/web-app/storybook}}

\section{Discussion and Conclusion}
\label{sec:discussion-and-conclusion}

The TIB AIssistant is not a direct alternative or replacement of existing LLM user interfaces. The AIssistant is tailored toward research and provides various advantages compared to regular LLM interfaces. First and foremost, the integration of assistants provides researchers with a library of curated prompts, which enables them to use LLMs within their research without the need to craft these prompts themselves. LLMs themselves provide almost limitless possibilities for research, but without curated lists of use cases (i.e., prompts), for many researchers, these benefits remain out of reach. Furthermore, the integrated tool library focuses on research-oriented tools and services, which are not typically provided by regular LLM interfaces. Tools can utilize dedicated user interface components, via the generative UI paradigm, to offer additional interaction possibilities beyond the conversational (i.e., text-only) approach of regular LLM interfaces. In the end, we envision the AIssistant as a platform that integrates many different prompts, models, and tools from a wide variety of research domains, created and maintained by the research communities themselves. Alongside the open source nature of the AIssistant, the collaboration of research communities to support different domains is a key aspect of our approach. This sets the AIssistant apart from existing approaches and other commercial initiatives. To facilitate community collaboration, we have provided developer documentation and an easy-to-use format for defining assistants to encourage community contributions. At the current stage, we have provided a framework in which the previously mentioned aspects can be integrated. We demonstrated how different assistants and prompts can be used to perform research tasks, however, the actual implementation of a large-scale domain-specific prompt and tool library is left for future work. To summarize, although the individual tasks within the assistant can indeed also be executed by existing LLM interfaces, the AIssistant platform is tailored toward scholarly research. The integration of different scholarly tools and domain-specific prompts into a single platform makes it possible for any researcher, regardless of the AI proficiency, to leverage AI in their workflows. 

In the current implementation, all assistants use the same model (GPT 4o-mini). The technology stack supports switching models and providers. For now, we do not want to be limited by the capabilities of the model, e.g., the ability to select the right tool for tool calling, or the capability to generate research ideas for arbitrary domains. However, we envision that assistants will use a model tailored to the task at hand. This means that smaller, locally hosted, or fine-tuned models are most likely viable options for various assistants. Using such approaches reduces the models' required resources, benefiting both costs and environmental impact, and is therefore an important research direction for future work. Another aspect of future work is the evaluation of our approach. This will focus both on specific assistants and how well they are able to accomplish their respective objectives, as well as evaluating the platform from a usability perspective. Experimenting with these smaller models is out of scope for this work. 

To conclude, in this work, we demonstrated what an AI-supported research assistant looks like, implemented in the TIB AIssistant. With this, we lay the foundation for a comprehensive community-maintained platform for researchers who want to leverage AI for their work. We demonstrated how different assistants work together by storing intermediary results in the assets storage. Also, we showed how external tools play an essential role in providing a research-oriented platform. 
Finally, we used a RO-Crate export functionality to provide machine-readable provenance data of the generated content. In the future, we plan to further simplify the process of adding assistants and tools to foster further community involvement. To accomplish this, we are experimenting with no-code solutions, where graphical user interfaces allow users to configure the platform further. Additionally, we plan to experiment with different models and add additional assistants and tools to support more complete research life cycles for various domains.  

\begin{acknowledgments}
We thank our colleague Mohamad Yaser Jaradeh for his valuable comments in reviewing this paper Additionally, we want to thank the entire AIssistant team for their contributions to the platform, including research and development efforts. This work was co-funded by NFDI4DataScience (ID: 460234259) and by the TIB Leibniz Information Centre for Science and Technology.
\end{acknowledgments}

\section*{Declaration on Generative AI}
  During the preparation of this work, the authors used ChatGPT and Grammarly in order to: Paraphrase and reword, improve writing style, and Grammar and spelling check. Also, the authors used OpenAI TTS for generating the voice-over in the demo video. After using these services, the authors reviewed and edited the content as needed and take full responsibility for the publication’s content. 

\bibliography{refs}

\end{document}